\documentclass[conference]{IEEEtran}
\IEEEoverridecommandlockouts
\usepackage{cite}
\usepackage{amsmath,amssymb,amsfonts}
\usepackage{algorithmic}
\usepackage{graphicx}
\usepackage{dsfont}
\usepackage{textcomp}
\usepackage{xcolor}
\def\BibTeX{{\rm B\kern-.05em{\sc i\kern-.025em b}\kern-.08em
    T\kern-.1667em\lower.7ex\hbox{E}\kern-.125emX}}
\begin{document}

\newtheorem{definition}{\textbf{Def.}}
\newtheorem{theo}{\textbf{Theorem}}

\title{A Novel Information-Driven Strategy for Optimal Regression Assessment
}

\author{\IEEEauthorblockN{Benjamín Castro\textsuperscript{1}, Camilo Ramírez\textsuperscript{1}, Sebastián Espinosa\textsuperscript{1}, Jorge F. Silva\textsuperscript{1}, Marcos E. Orchard\textsuperscript{1}, Heraldo Rozas\textsuperscript{1}}
\IEEEauthorblockA{\textit{\textsuperscript{1}Information and Decision Systems Group, Department of Electrical Engineering} \\
\textit{Universidad de Chile}\\
Santiago, Chile \\
benjamincastro@ug.uchile.cl}}

\maketitle

\begin{abstract}
In Machine Learning (ML), a regression algorithm aims to minimize a loss function based on data. An assessment method in this context seeks to quantify the discrepancy between the optimal response for an input-output system and the estimate produced by a learned predictive model (the student). Evaluating the quality of a learned regressor remains challenging without access to the true data-generating mechanism, as no data-driven assessment method can ensure the achievability of global optimality. This work introduces the Information Teacher, a novel data-driven framework for evaluating regression algorithms with formal performance guarantees to assess global optimality. Our novel approach builds on estimating the Shannon mutual information (MI) between the input variables and the residuals and applies to a broad class of additive noise models. Through numerical experiments, we confirm that the Information Teacher is capable of detecting global optimality, which is aligned with the condition of zero estimation error with respect to the -- inaccessible, in practice -- true model, working as a surrogate measure of the ground truth assessment loss and offering a principled alternative to conventional empirical performance metrics.
\end{abstract}

\begin{IEEEkeywords}
regression, assessment, mutual information
\end{IEEEkeywords}

\section{Introduction}
The regression task in Machine Learning (ML) seeks to capture the relationship between a dependent variable $Y \in \mathbb{R}^q$ (often referred to as the response or output) and an independent variable $X \in \mathbb{R}^p$ (explanatory variable or input) expressed by a joint distribution $P_{X,Y}$\cite{weisberg2005applied,montgomery2012introduction,hastie2009elements,green1993nonparametric}. 
Formally, the regression task involves identifying a function within a hypothesis space that minimizes the discrepancy of the estimation of the response variable with respect to its actual value. The performance of a regression model can be evaluated under various loss functions, among which the mean squared error (MSE) is one of the most commonly used. 

In ML, regression models are built from data-driven approaches where consistency---i.e., asymptotic convergence to the optimal regressor based on the joint distribution $P_{X,Y}$ from i.i.d.\ samples---is a desirable property. Classical methods approximate this optimal regressor by minimizing empirical loss, often via gradient descent. Parametric models, such as linear regression or multilayer perceptrons (MLPs), constrain the hypothesis space to predefined functional forms, offering computational tractability and interpretability but limited flexibility to capture the true data-generating process or guarantee convergence to the minimum MSE solution \cite{HORNIK1989359, convergence_rates}. Non-parametric methods---e.g., tree-based models or $k$-nearest neighbors---seek to approximate the true regression function while balancing accuracy and regularization \cite{xgboost, reg_trees}.

Key challenges in regression include: \textbf{(i)} limited data availability, especially for complex models like DNNs \cite{challenges_nn}, a critical issue in fields such as astronomy \cite{astro_data}, business \cite{ecommerce, ecommerce2}, and healthcare \cite{healthcare}; \textbf{(ii)} convergence issues in empirical MSE minimization due to local minima \cite{loss_surface, petzka_non-attracting_2021}; and \textbf{(iii)} selecting a hypothesis space that is expressive yet computationally feasible \cite{model_cmplx}.

Performance evaluation is typically empirical, using a validation set to estimate the MSE and guide model selection or hyperparameter tuning \cite{cv_classic, cv_model_selection}. While effective for local comparisons, this approach cannot determine if a model has reached \emph{global optimality}. Assessing this from data alone would be valuable for meta-learning, early stopping, and performance guarantees. However, such validation is intractable: theoretical bounds like the conditional variance require knowledge of $P_{X,Y}$, and even with this, computing Cramér-Rao bounds or minimax risks in high dimensions is often infeasible \cite{lehmann1998theory,van2000asymptotic, wasserman2006all, belkin2019reconciling}.

This motivates a practical, model-agnostic method to detect global optimality without prior knowledge of the ground-truth model. We address this with a novel information-theoretic approach offering strong guarantees under minimal assumptions.

\subsection{Contributions}
We propose the \textbf{Information Teacher}, a decision rule based on estimating the Shannon Mutual Information (MI) between inputs and residuals. Our method provides the first asymptotic and finite-sample guarantees for detecting global MSE optimality under a broad family of additive noise models. Its strengths are: \textbf{(1)} a necessary and sufficient MI-based condition for global optimality; \textbf{(2)} consistency and finite-sample guarantees; and \textbf{(3)} distribution-free detection without assumptions on input or residual distributions. Controlled experiments confirm the ability to detect MMSE optimality from data alone, consistent with our theory.

\subsection{Related Work}
Optimality in regression has been studied through lower bounds on estimation error, such as the Cramér-Rao lower bound \cite{lehmann1998theory,van2000asymptotic}, valid under strong regularity and likelihood knowledge \cite{keramat2019cramer, cr_boun_batteries, he2024cramer}. In high-dimensional or nonparametric settings, minimax risk offers a distribution-free characterization \cite{tsybakov2008introduction, donoho1994minimax}, though still impractical for finite-sample validation.

Recently, validation methods for assessing regression strategies were proposed to give an accurate insight into the algorithm's performance relative to a true model \cite{error_bounds}. This work assesses global optimality by providing an error bound under the strong assumption that the input-output system is a Gaussian process \cite{error_bounds}. In a related direction, recent works have explored adaptive estimation strategies capable of approaching optimal risk without relying on full knowledge of the model structure. These include procedures that adapt to local regularity conditions or unknown smoothness of the target function, offering near-optimal performance under weaker assumptions \cite{juditsky2009nonparametric}. This has led to renewed interest in non-asymptotic risk bounds for the estimator’s performance with finite samples without requiring consistency or unbiasedness \cite{wainwright2019high}. 

Finally, in the context of MSE risk minimization, it is worth emphasizing that no evaluation scheme based solely on data is capable of assessing global optimality of a regression algorithm without oracle knowledge of the true underlying model. As a result, determining whether a given algorithm achieves the lowest possible MSE between the response and the prediction remains an open challenge \cite{donoho2000high,wainwright2019high}.

%\subsection{Organization}
The rest of the paper is organized as follows. Section \ref{sec:preliminaries} introduces the concept of the \emph{teacher} as an assessment agent, along with the learning context of the problem. In Section \ref{sec_info_teacher}, we present and motivate our method, accompanied by a formal performance analysis (Theorem \ref{theo:TheoremProps}). Section \ref{sec_experiments} provides empirical validation that demonstrates and confirms the effectiveness of our method. Finally, we discuss the practical advantages of implementing our method in real-world scenarios and outline potential directions for future work. 

\section{Preliminaries}
\label{sec:preliminaries}

In this section, we introduce the preliminary concepts used in the context of regression. Let $(X, Y)$ be a pair of random vectors taking values in $\mathcal{X} \times \mathcal{Y} \subset \mathbb{R}^p \times \mathbb{R}^q$, where $X$ is referred to as the input variable and $Y$ as the output variable. The marginal distribution of $X$ is denoted by $P_X \in \mathcal{P}(\mathcal{X})$, where $\mathcal{P}(\mathcal{X})$ is the set of all probability distributions over $\mathcal{X}$.  

A learning algorithm of size $n$, denoted $\mathcal{A}_n (\cdot)$, is defined as a function $ \mathcal{A}_n : (\mathcal{X} \times \mathcal{Y})^n \to \mathcal{Y}^\mathcal{X}$, that maps a dataset of $n$ i.i.d.\! pairs $(x_i, y_i)_{i=1}^n$ to a function $\hat{f}(\cdot) \in \mathcal{F} \subset \mathcal{Y}^\mathcal{X}$, where $\mathcal{F}$ denotes the hypothesis space %. Then, a learning algorithm is a general class of algorithms designed to learn patterns from data 
\cite{bousquet2002stability, bousquet2020sharper}.
Regression algorithms are a specific subclass of learning algorithms designed to predict continuous variables. Formally, their goal is to learn a function $\hat{f} : \mathcal{X} \to \mathcal{Y}$ that maps an input $X \in \mathcal{X}$ to a real-valued (or vector-valued) output $Y \in \mathcal{Y} \subset \mathbb{R}^q$. Unlike classification algorithms, which assign discrete labels to inputs, regression models yield numerical estimates that can take values from an uncountably infinite set, making them especially suitable for forecasting, curve fitting, and other tasks involving continuous target spaces.

Regression algorithms are designed to minimize a predefined loss (or risk) that quantifies the discrepancy between the true outputs and the predicted values. In this work, we adopt the squared loss function %--Least Squares context-- 
\cite{regression_book_} given by
\begin{equation}
    \ell : \mathcal{X} \times \mathcal{Y} \times \mathcal{Y}^\mathcal{X} \to \mathbb{R}^{+}, \quad \ell(X, Y, \hat{f}) = (Y - \hat{f}(X))^2.
\end{equation}

The learning process aims to find the hypothesis $\hat{f}(\cdot) \in \mathcal{F}$ that minimizes the expected risk, given by 
\begin{equation}
\mathbb{E}_{X,Y}[\ell(X, Y, \hat{f})] = \mathbb{E}_{X,Y}[(Y - \hat{f}(X))^2].
\end{equation}

It is well established that the function that minimizes the expected squared loss (or mean square error (MSE)) is the conditional expectation of $Y$ given $X$, that is, $f(x) = \mathbb{E}[Y \mid X=x]$ \cite{pml1Book}.

A richly adopted assumption in regression problems is that the response variable can be expressed as the sum of a deterministic component and a random perturbation. This leads to the widely adopted class of \emph{additive noise models} \cite{roberts1987digital, kay1993fundamentals, wheelwright1998forecasting}, which provide a flexible and interpretable framework for modeling uncertainty and variability in real-world scenarios \cite{draper1998applied}. In particular, we assume that the probabilistic relationship between input and output takes the form of 
\begin{equation}
    Y = f(X) + W,
\end{equation}
where $f(\cdot) \in \mathcal{Y}^\mathcal{X}$ is a measurable function (the target regression function), and $W \sim P$ is a zero-mean noise that is independent of $X$. In this context, the noise $W$ can be viewed as an adversarial component that perturbs the true signal $f(X)$. 

%Observe that this formulation also aligns naturally with the objective of regression under the squared loss: to recover a function $\hat{f}$ that minimizes the variance of the residual noise, i.e., to make the discrepancy $Y - \hat{f}(X)$ as small as possible in expectation.

In the following section, we formally introduce the additive model framework and discuss its implications for learning and estimation.

\subsection{Additive Noise Models}
Additive noise models constitute a widely used and theoretically well-grounded framework for describing the relationship between inputs and outputs. They assume that the response $Y$ can be expressed as the sum of a deterministic component $f(X)$ and a stochastic term $h(W)$, independent of $X$. This structure captures a broad range of practical settings while enabling tractable analysis and provable results for optimality detection. In this work, we adopt the family of additive noise models as formally introduced by \cite{ramirez}.

\definition{}: A pair $(X,Y)$ is said to follow an additive noise model, denoted by $(X,Y) \sim \textup{add}(f, h; P_X)$, if both conditions hold:
\begin{itemize}
    \item $\mathbb{P}(X \in A) = P_X(A), \quad \forall A \in \mathcal{B}(\mathbb{R}^p)$.
    \item $\mathbb{P}\big( (X,Y) = (X, f(X) + h(W)) \big) = 1$.
\end{itemize}
Here, $P_X$ is the marginal distribution of $X$, $W \sim U([0,1])$ is independent of $X$, and $f:\mathbb{R}^p \to \mathbb{R}^q$ and $h:[0,1] \to \mathbb{R}^q$ are measurable functions. This is a particular case of the Transfer Theorem \cite[Chapter~8, Theorem~8.17]{transfer}.

While this family does not encompass all possible noise structures—e.g., multiplicative or high variable noise—it strikes a balance between generality and analytical tractability. Importantly, many nonlinear regression problems can be recast into an additive form (possibly after transformations), making this framework sufficiently rich for our purposes while allowing sharp theoretical guarantees. For example, the Kolmogorov–Arnold representation theorem theoretically ensures that any continuous multivariate function can be decomposed into sums and compositions of univariate functions—justifying, in principle, the approximability of complex nonlinearities through additive structures \cite{kolmogorov1957representation}

\subsection{Shannon Mutual Information}

The mutual information (MI) between two random variables was introduced by {\em Shannon} in 1948 \cite{shannon}, and it is widely used to quantify high-order statistical dependencies \cite{cover}. Given two continuous random variables $X \in \mathcal{X}$ and $Y \in \mathcal{Y}$, the MI is expressed as follows \cite{kolmogorov}:
\begin{equation}
    \mathcal{I}(X;Y) = \int_{\mathcal{X}} \int_{\mathcal{Y}} f_{X,Y} (x,y) \cdot \log \left (\frac{f_{X,Y} (x,y)}{f_X (x) \cdot f_Y(y)}\right) \mathrm{d}x\:\mathrm{d}y,
    \label{eq:kolmogorov_mi}
\end{equation}
where $f_{X,Y}(\cdot, \cdot)$ corresponds to the joint probability density function (pdf) of $(X,Y)$, and $f_X(\cdot)$ and $f_Y(\cdot)$ are the marginal pdfs of $X$ and $Y$, respectively. An important property is that MI between $X$ and $Y$ is $0$ if, and only if, both random variables are independent \cite{cover}. 

The mutual information offers two distinctive advantages for our problem. First, MI is a well-established quantity %in information theory 
that captures all forms of statistical dependence, beyond linear and low-order dependencies. Second, MI has a clear operational interpretation for regression %: it quantifies the amount of information about $X$ that remains in $Y$. In regression settings, 
when analyzing the residual $R := Y - \hat{f}(X)$, MI measures how much information about $X$ is still present in $R$, and thus how far $\hat{f}(\cdot)$ is from achieving the conditional expectation $\mathbb{E}[Y|X=x]$.

% Despite its theoretical appeal, estimating mutual information in a high-dimensional, nonparametric setting presents significant technical challenges.

In the next section, we will introduce the concept of \textit{teacher} in regression, which could be defined as an agent that qualifies the performance of a given algorithm (the \textit{student}) through a decision rule, to assess the response of the \textit{student} based on its knowledge of the task.

\subsection{The Teacher: A Decision Agent for Regression Algorithms}
We frame our problem as an interaction between a \emph{student}—a regression algorithm $\mathcal{A}_n(\cdot)$—and a \emph{teacher}, an external agent that evaluates the quality of the learned predictor $\hat{f}(\cdot)$ from data $\mathcal{D} = \{(x_i, y_i)\}_{i=1}^n$. The teacher’s goal is to decide, according to a criterion $S \subset \mathcal{Y}^\mathcal{X}$, whether $\hat{f}$ meets a target property (approve) or not (reject). Formally:

\begin{equation}
\label{eq:teacher_original}
    T(\hat{f}) =  \left \{ \begin{array}{ll}
         1 &  \textup{if} \quad \hat{f}\in S,\\
         0 & \textup{if} \quad \hat{f} \in S^c.
    \end{array} \right.
\end{equation}

Ideally, our goal is to define \textit{teachers} based on criteria that characterize when an estimator $\hat{f}(\cdot)$ qualifies %as an \textit{Intelligent Student} 
—that is, when it corresponds to the \textit{answer} minimizing the MSE---, i.e., when $\hat{f}(\cdot)$ achieves the minimum mean squared error (MMSE). In the context of regression, %particularly within the \textit{Model Selection} framework, 
such criteria $S$ are often approximated by a simple threshold $a \in \mathbb{R}^{+}$ on the MSE of the \textit{answer} in the following form:
\begin{equation}
\label{eq:naive_teacher}
    T_N(\hat{f}, P_{X,Y}) =  \left \{ \begin{array}{ll}
         1 &  \textup{if} \quad \mathbb{E}_{X,Y} \{ (Y-\hat{f}(X))^2\}\leq a,\\
         0 & \textup{if} \quad \mathbb{E}_{X,Y} \{ (Y-\hat{f}(X))^2\}\ > a.
    \end{array} \right.
\end{equation}
The threshold $a>0$ denotes an implicit rule.  %, often corresponding to a lower bound provided by an estimation through Cross-Validation or similar methods. 
Notice that this definition requires the true joint distribution, nevertheless, the value of the MSE  can be estimated in a data-driven way from an empirical estimation of $\mathbb{E}_{X,Y} \{ (Y-\hat{f}(X))^2\}$ using a validation set (i.i.d.\! samples of $P_{X,Y}$. 
Importantly, in the approach  in Eq. (\ref{eq:naive_teacher}) there is no guarantees that the threshold $a$ matches the global optimum $a^* = \mathbb{E}_{X}\{\textup{Var}(Y\mid X)\}$ as this is a function of the unknown model $P_{X,Y}$. 
%In practice, this optimal value is often unattainable due to the inherent limitations of the hypothesis space to which the \textit{student}'s predictions are constrained. 
Consequently, this approach will be sub-optimal for the task of assessing global optimality. 

\subsection{The Oracle Teacher}
It is convenient for our analysis to consider an oracle teacher. This is an assessment method that has access to the true model $f(\cdot)$, i.e., the MMSE solution of the problem, to compare with $\hat{f}(\cdot)$. In this scenario, the oracle criterion 
%defining the \textit{Intelligent Student} condition from a ground-truth perspective 
corresponds to:   
%the \textit{answer} that achieves exactly zero MSE with respect to the model $f$ under the marginal distribution $P_X$, formally defined as follows.
\definition{}:
\label{Oracleteacher}
For $ (X,Y) \sim P_{X,Y} \in \mathcal{P} (\mathcal{X} \times \mathcal{Y} )$ and $P_X \in \mathcal{P}(\mathcal{X})$ we define the oracle teacher $\mathbf{T}_{\mathrm{O}}: \mathcal{Y}^{\mathcal{X}} \times \mathcal{F} \times\mathcal{P}( \mathcal{X}) \to \{0,1\}$ as
\begin{equation}
\label{eq:ot}
    \mathbf{T}_{\mathrm{O}}(f, \hat{f}, P_X) =  \left \{ \begin{array}{ll}
         1 &  \textup{if} \quad \mathbb{E}_X \{ ( f(X) - \hat{f} (X) )^2\} = 0, \\
         0 & \sim.
    \end{array} \right.
\end{equation}
This oracle represents the gold standard of assessment: it directly checks whether $\hat{f}$ matches $f$ pointwise under $P_X$. Unfortunately, in real-world problems, $f$ is unknown, making this ideal test infeasible.

The challenge, therefore, is to design a \emph{practical teacher}—one that, without access to $f$ or $P_{X,Y}$, can still determine whether $\hat{f}$ has reached the MMSE solution. In the next section, we present the \emph{Information Teacher}, an information-theoretic criterion that approximates the oracle's behavior using only data.

\section{The Information Teacher}
\label{sec_info_teacher}
Leveraging the structure of additive noise models, we propose the following decision rule based on measuring the amount of mutual information between the input $X$ and the residual $R = Y - \hat{f}(X) $. 

\definition{}:
\label{InfoTeacher}
For $ (X,Y) \sim P_{X,Y} \in \mathcal{P} (\mathcal{X} \times \mathcal{Y} )$ we define the information teacher as a binary function $ \mathbf{T}_{\mathrm{I}}: \mathcal{F} \times \mathcal{P}(\mathcal{X}\times\mathcal{Y})\to \{0,1\}$ expressed by
\begin{equation}
\label{eq:ito}
    \mathbf{T}_{\mathrm{I}}(\hat{f}, P_{X,Y}) =  \left \{ \begin{array}{ll}
         1 &  \textup{if} \quad \mathcal{I} (X; Y-\hat{f}(X)) = 0,\\
         0 & \sim.
    \end{array} \right.
\end{equation}
    Intuitively, if an estimator $\hat{f}(\cdot)$ successfully recovers the true function $f(\cdot)$, the residual $R$ should be statistically independent of $X$. This motivates the use of MI --~which is $0$ if, and only, if there are no statistical dependencies \cite{cover} ~-- as a criterion to detect deviations from independence, and thereby to assess the optimality of the estimator. 
    Importantly, the formal  connection between this information-driven approach, which does not  use explicit information of the true model $P_{X,Y}$, and the oracle teacher in Eq.(\ref{eq:ot}) is presented in the following result:  
    %Yet, Eq. (\ref{eq:ito}) still imposes a strong assumption, as the true generative model is not always accessible. However, in the particular case of additive noise models, we instead introduce a necessary and sufficient condition to evaluate global optimality for the \textit{answer} that satisfies the \textit{Intelligent Student} condition.

\theo{} \label{theo:teo1} \cite[Th.1]{ramirez}: Let $X$ and $Y$ be random variables taking values in $\mathbb{R}^p$ and $\mathbb{R}^q$, respectively, such that $(X,Y) \sim \textup{add}(f,h;P_X)$ and let $\hat{f}:\mathbb{R}^p\to \mathbb{R}^q$ be a measurable function, then 

\begin{equation}
   \mathbb{P} \left( \hat{f}(X) = f(X)\right) = 1 \Leftrightarrow \mathcal{I}(X; Y -\hat{f}(X)) = 0.
\end{equation}

Theorem \ref{theo:teo1} offers a connection between the idea of detecting statistical independence with the almost-surely equivalence between the \textit{answer} $\hat{f}$ and the unknown true model $f$ under the marginal input distribution $P_X$. This offers a powerful direction where the oracle assessment in Eq.(\ref{eq:ot})  could be implemented with only the input variable $X$ and the residual $Y-\hat{f}(X)$. Then, our Information Teacher (Def. \ref{InfoTeacher}) can implement the  Oracle Teacher (Def. \ref{Oracleteacher}), in terms of assessing global optimality in the MMSE sense. % (\textit{Intelligent Student} condition).

While Eq.~(\ref{eq:ito}) avoids the explicit use of the optimal unknown function $f(\cdot)$, it implicitly requires the joint distribution to compute MI. To address this practical limitation, we will use a validation set, traditionally available for the assessment in machine learning, to implement a distribution-free estimation of the MI and consequently a practical data-driven version of the information teacher in  Eq.(\ref{eq:ito}).  
%In this sense, acquiring samples from the joint distribution may be more attainable than having analytical access to the true model $f$. 
In particular, we propose a data-driven version of the Information Teacher:% in the following definition.

\definition{}: For $ (X,Y) \sim P_{X,Y} \in \mathcal{P} (\mathcal{X} \times \mathcal{Y} )$, we define the data-driven information teacher $\hat{\mathbf{T}}_m : \mathcal{F} \times (\mathcal{X} \times \mathcal{Y})^m \to \{0,1\}$, $\mathcal{D}_m \in (\mathcal{X} \times \mathcal{Y})^m$ as
\begin{equation}
\label{eq:ddit}
    \hat{\mathbf{T}}_m(\hat{f}; \mathcal{D}_m) \ =  \left \{ \begin{array}{ll}
         1 &  \textup{if} \quad \hat{\mathcal{I}}_m (X_{\mathcal{D}_m}; Y_{\mathcal{D}_m} - \hat{f}(X_{\mathcal{D}_m})) = 0,\\
         0 & \sim,
    \end{array} \right.
\end{equation}
where $\mathcal{D}_m$ denote $m$ i.i.d. realizations of $(X;Y)\sim P_{X,Y}$.

%This rule is the empirical version of the information teacher. 
For the data-driven estimation of the MI, i.e., $\hat{\mathcal{I}}_m (X_{\mathcal{D}_m}; Y_{\mathcal{D}_m} - \hat{f}(X_{\mathcal{D}_m}))$ in (\ref{eq:ddit}), which is a central element of our scheme,  we adopt the algorithm proposed in \cite{tsp}. This estimator has state-of-the-art convergence properties for detecting the condition of zero MI (independence)  and is a consistent estimator of the MI distribution-free. 
 Importantly, this estimator will be instrumental for detecting the mismatch enunciated in Theorem~\ref{theo:teo1}. The following section introduces theoretical guarantees of our data-driven Information Teacher.

\subsection{Asymptotic and Finite-Length Properties}
The following Theorem presents the main performance results of our method.  
\theo{} \label{theo:TheoremProps}
Let $(X,Y)\sim \textup{add}(f,h;P_X)$ and $\mathcal{D}_m$ be $m$ i.i.d. samples of $(X,Y) \sim \textup{add}(f,h;P_X)$. Let $\hat{f}:\mathbb{R}^p\rightarrow\mathbb{R}^q$ denote the output of  the \emph{data-driven information teacher} -- i.e. $\hat{\mathbf{T}}_m(\cdot;\cdot)$ in Eq.(\ref{eq:ddit}). Then,  the following properties hold: 
\begin{itemize}
    \item[(i)] Strong consistency:
    \begin{align}
        \text{If } f(X) \overset{\textup{a.s.}}{=} \hat{f}(X) &\Rightarrow \hat{\mathbf{T}}(\hat{f};\mathcal{D}_m)\xrightarrow[m\rightarrow\infty]{\text{a.s.}}1,\\
        \text{If } f(X) \overset{\textup{a.s.}}{\neq} \hat{f}(X) &\Rightarrow \hat{\mathbf{T}}(\hat{f};\mathcal{D}_m)\xrightarrow[m\rightarrow\infty]{\text{a.s.}}0,
    \end{align}
    where ${f(X) \overset{\textup{a.s.}}{=} \hat{f}(X)}$ and ${f(X) \overset{\textup{a.s.}}{\neq} \hat{f}(X)}$ are shorthand notations for the condition ${\mathbb{P}(f(X) = \hat{f}(X)) = 1}$ and ${\mathbb{P}(f(X) = \hat{f}(X)) < 1}$, respectively, and ${\hat{\mathbf{T}}_m(\hat{f};\mathcal{D}_m)\xrightarrow[m\rightarrow\infty]{\text{a.s.}} k}$ is a notation for almost-surely convergence of the random decisions to  $k$~---~$k\in\{0,1\}$.
    \item[(ii)] Exponentially-fast detection on the global optimum:
    \begin{align}
       \text{If }  f(X) \overset{\textup{a.s.}}{=} \hat{f}(X)\Rightarrow\:&\mathbb{P}(\sup\{k:\hat{\mathbf{T}}_{k}(\hat{f};\mathcal{D}_k)=0\}\geq m) \\
        & \leq K\exp(-m^{1/3}),\forall m\in\mathbb{N},
    \end{align}
    with $K>0$ a universal (distribution-free) constant.
    \item[(iii)] Error convergence to zero:
    \begin{align}
    \text{If } f(X)\overset{\textup{a.s.}}{=}\hat{f}(X) \Rightarrow\:&\mathbb{P}(\hat{\mathbf{T}}_m(\hat{f};\mathcal{D}_m)=0) \\
     & \leq K\exp(-m^{1/3}),\forall m\in\mathbb{N},\\
    \text{If } f(X)\overset{\textup{a.s.}}{\neq}\hat{f}(X)\Rightarrow\:&\lim_{m\rightarrow\infty}\mathbb{P}(\hat{\mathbf{T}}_m(\hat{f};\mathcal{D}_m)=1) = 0,
    \end{align}
    with $K>0$, a universal (distribution-free) constant.
\end{itemize}
The proof of this three results are presented in the Supplemental Material.

First, it is important to note that the condition $f(X)\overset{\textup{a.s.}}{=}\hat{f}(X) $ means that the regressor $\hat{f}(\cdot)$ is optimal because this is equivalent to state that $\mathbb{E}_X \{ ( f(X) - \hat{f} (X) )^2\} = 0$ and, consequently, that $\hat{f}(\cdot)$ achieves the MMSE for the true model $(X,Y)\sim \textup{add}(f,h;P_X)$.  
Given this context, Theorem 1 shows our  method's asymptotic and non-asymptotic performance  when using validation data as its unique source to perform an binary assessment for optimality. 
Point {\bf (i)} proves that our method convergences to the right assessment  by increasing the validation set's sampling size.. Point {\bf (ii)} proves that the probability that our teacher yields any wrong decision after a fixed number of samples vanishes exponentially fast with $m$. Finally, Point {\bf (iii)} shows how the error of a wrong assessment converges to zero. To our knowledge, no other data-driven methods in the literature have demonstrated this capacities to achieve the right assessment from data.

\section{Experimental Setting}
\label{sec_experiments}
In this section, we provide numerical results that support the Information Teacher capabilities of discriminating local and global optimality stated in Theorem \ref{theo:TheoremProps}. The Information Teacher works as a practical diagnostic tool, which could be understood as a surrogate of the Oracle Teacher metric \eqref{eq:ot} ($\mathbb{E}_X\{(f(X) - \hat{f}(X))^2 \}$).  This implies  a powerful advantage in real-world contexts, where the true data-generation model is not accessible. We assess the effectiveness of our method through two complementary experimental settings. First, we test its capability of detecting optimality in a controlled synthetic scenario, where we leverage full knowledge of the ground-truth distribution to evaluate the true diagnostic power of our method and compare it with the classical validation-MSE baseline. Then, to further demonstrate the practical utility of our approach, we also evaluate our method on a real-world dataset from a combined-cycle power plant, involving nearly 10,000 records with environmental and operational features. This allows us to test our Information Teacher in a setting where the true optimum is unknown, thus reflecting more realistic conditions.

%In the synthetical data experiments, we present two main cases: the first one (with favorable conditions) where the algorithm achieves the global optimum, contrasted with zero ground-truth squared error; and a second unfavorable case, with an algorithm that is forced to achieve local optima due to the parametric space of the selected algorithm.
\subsection{Synthetic Data}
We evaluate the \emph{Information Teacher} on a controlled continuous input–output system to test its ability to detect both global and local MSE optima in regression. The ground truth is $f(X) = \sin(10X) + \omega,\quad \omega \sim \mathcal{N}(0,0.25),$ which allows full control over the learning task and access to the true function.

Two scenarios are considered. In the \textit{favorable case}, the hypothesis space is sufficiently rich: an MLP with two hidden layers of $128$ units, trained with Adam \cite{adam}, reaches the global optimum. In the \textit{unfavorable case}, the space is deliberately limited: a single hidden layer of $32$ units trained with SGD, prone to suboptimal convergence. For reference, we also compute the oracle squared error and the validation MSE—quantities unavailable in real datasets—to compare our method with the ideal (oracle) and classical validation-based assessments

\begin{figure*}[h]
    \centering
    \includegraphics[width=0.5\linewidth]{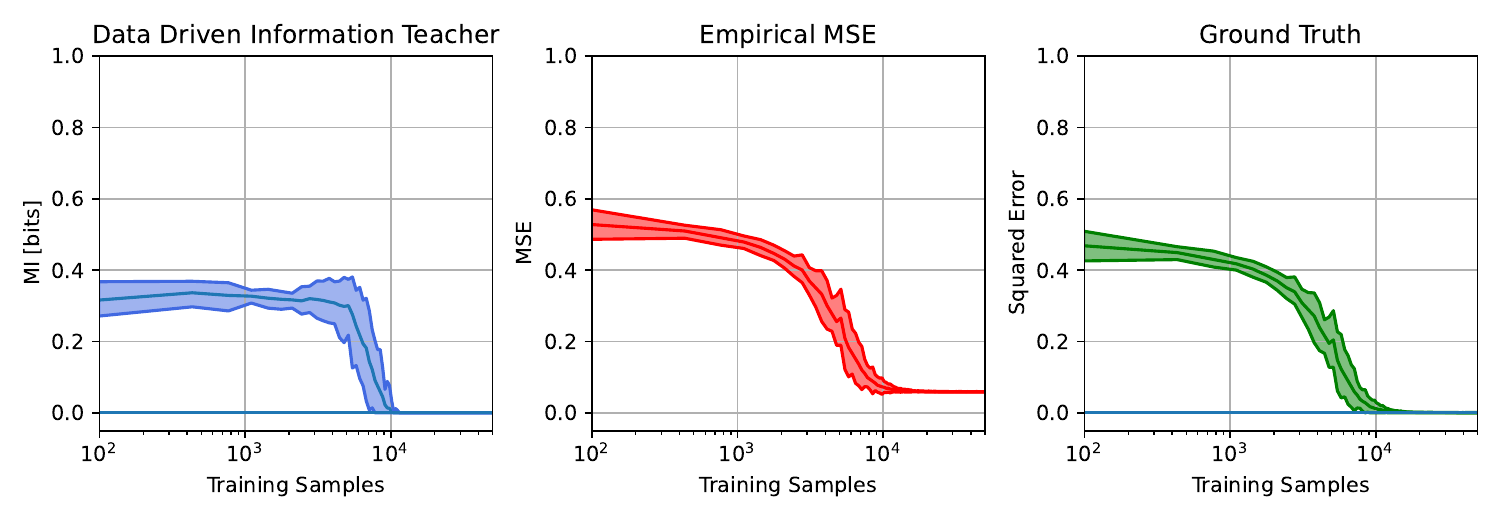}
    \caption{Performance of \textit{favorable case} (median and IQR intervals)}
    \label{fig:test_favor}
\end{figure*}

\begin{figure*}[h]
    \centering
    \includegraphics[width=0.5\linewidth]{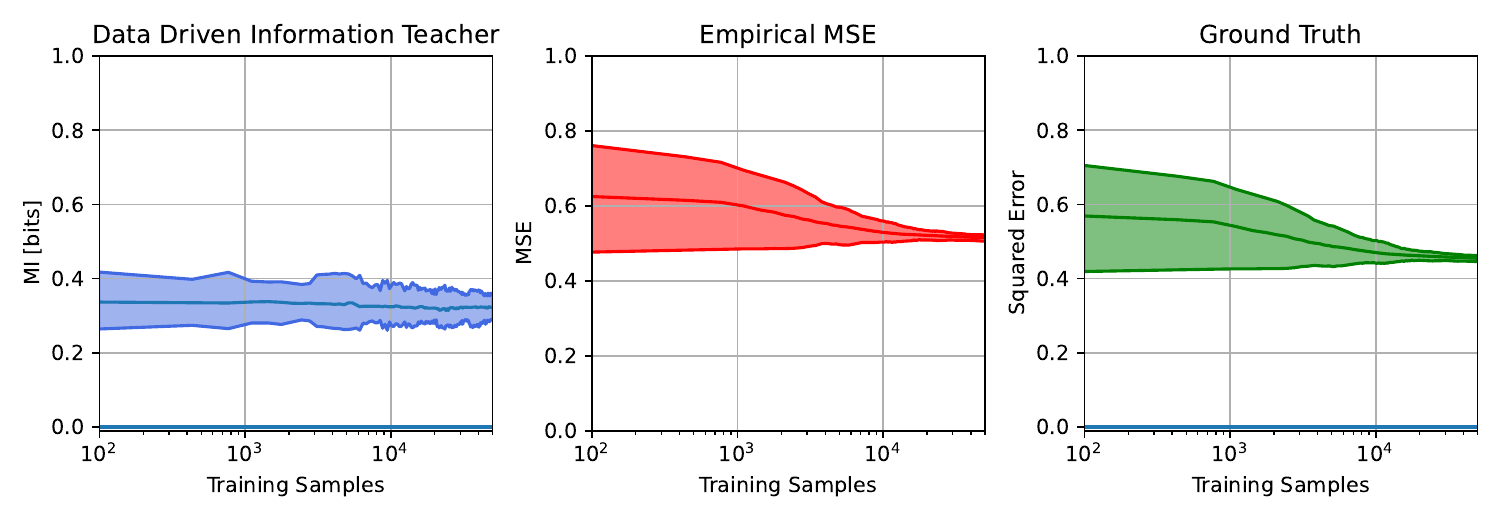}
    \caption{Performance of \textit{unfavorable case} (median and IQR intervals)}
    \label{fig:test_unfavor}
\end{figure*}

Figures \ref{fig:test_favor} and \ref{fig:test_unfavor} show the distributional evolution of the computed metrics across different runs, characterized by their median and inter-quartile range (IQR). They illustrate how the Information Teacher metric evolves with the training sample size, which varies logarithmically from 100 to 50,000 samples. The learning rate and training batch size are fixed at $10^{-4}$ and 64 samples, respectively, for all experiments. An early stopping strategy is employed to prevent overfitting, using a tolerance of $10^{-4}$ in the validation loss and a patience of 5 epochs, with a maximum of 50 training epochs. Each experimental condition is repeated with 20 different random seeds, accounting for variability in both data sampling and model initialization. For evaluation, the Information Teacher metric, MSE, and the oracle metric (true squared error) are computed on a fixed i.i.d.\! validation set of 2,000 samples.

We see that in the \textit{favorable case} (Figure \ref{fig:test_favor}), the Information Teacher metric is aligned directly with the ground truth (oracle) squared error: both metrics collapse directly to zero (i.e., they evidence the reachability of the global optimum) on a training data size of approximately $10^4$ samples. This result supports the ability of the Information Teacher to work as a surrogate measure of the ground truth error (Oracle Teacher), without the need to access directly the true model and being a computationally feasible solution that only requires input and estimation residuals to make an accurate decision supported by Theorem \ref{theo:teo1}. In contrast, the empirical MSE (classic method) reaches a floating plateau whose optimality is not possible to assess without knowledge of the ground-truth distribution.
 

In contrast, the \textit{unfavorable case} (see Figure \ref{fig:test_unfavor}) shows the convergence of the MSE metric to a floating plateau value with low variance when the amount of training data is approximately $10^4$ samples. However, only with the information provided by the MSE metric, there are no guarantees to confirm whether the achieved point corresponds to a local or global optimum. Conversely, the Information Teacher metric converges to a non-zero value which implies its decision about the algorithm convergence to a local optimum; this is confirmed by the ground truth squared error, which also converge to a non-zero value. We remark that in a real-world scenario, the assessment of regression methods with the conventional MSE metric does not give any guarantee about the achievement of global optimality.

%Finally, these scenarios are chosen to reflect the trade-off between the training data availability and the algorithm performance, especially in cases where the data acquisition has associated costs, as mentioned in section 1 in contexts like marketing \cite{ecommerce, ecommerce2}, astronomy \cite{astro_data} and healthcare \cite{healthcare}, where the need of providing accurate predictions in the deployment of the algorithms plays a crucial role to achieve the desired tasks and the training resources are limited. Again, in real-world regression tasks, the true model is inaccessible. Hence, having a principled method for evaluating whether an algorithm has reached the global optimum --based only on available data and the learned predictions-- remains as a relevant and open problem \cite{data_scarcity}. Our approach aims to address this gap by providing a qualitative metric that can assess performance in terms of the training data quantity without requiring access to the true model.

\subsection{Real-world Data}
In this section, we present experiments performed in real-world data presented in \cite{real_world_data} and available in \cite{dataset_availability}. The data corresponds to 9,568 records of a combined-cycle electrical power plant, where the target variable is the value of the full load electrical power and the features are the ambient temperature, steam pressure, relative humidity, and atmospheric pressure. The experiments conducted on this dataset follow the same design as in the synthetic case and aim to validate the performance of the Information Teacher when assessing the trade-off between the amount of training data and the algorithm’s performance. However, unlike the synthetic case, it is not possible to perform a ground truth-based analysis due to the absence of an optimal reference.

The data split is performed with different random seeds for 20 random initializations, with a fixed validation set size of 2,000 samples; the training dataset size rises logarithmically from 100 to 7,000 samples. We train a shallow MLP regressor with a single hidden layer of 256 units, using an adaptive learning rate of $10^{-4}$ for the Adam optimizer, and we employ an early stopping criterion with the same characteristics as the synthetic experiment for a maximum of 300 epochs. Additionally, we perform a dimensionality reduction of the input features using Principal Component Analysis (PCA), from the original 4 features to 2 components, as a method to avoid redundancies. In contrast with synthetic experiments, we compute the RMSE measure instead of MSE to be consistent while comparing our results with the algorithms evaluated in \cite{real_world_data}.

\begin{figure*}[h]
    \centering
    \includegraphics[width=0.45\linewidth]{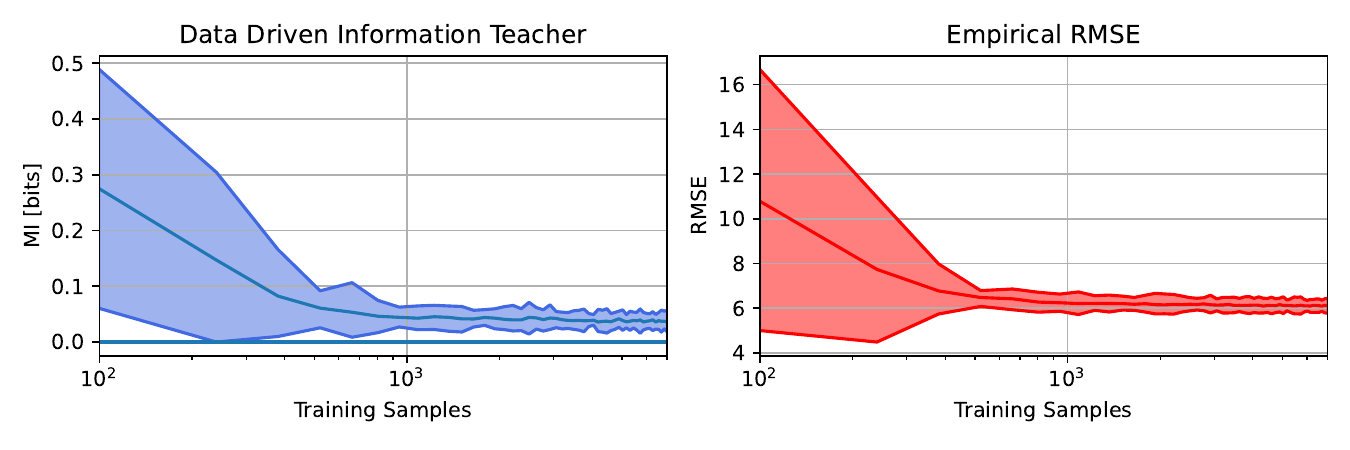}
    \caption{Performance of MLP Regressor (median and IQR intervals)}
    \label{fig:real-mlp}
\end{figure*}

Figure \ref{fig:real-mlp} presents the behavior of the Information Teacher metric in the real data scenario. The results show that the metric converges to a non-zero plateau with low variance, suggesting that the algorithm reaches a sub-optimal performance level. Notably, in the range between 100 and 
$10^3$ training samples, the Information Teacher indicates the attainment of independence, while the RMSE stabilizes around a value of 6—consistent with the best-performing regressor reported in \cite{real_world_data}. Again, the results remark the interpretability of the Information Teacher in a low-error regime, where the independence condition achievement is a necessary and sufficient condition for global MSE optimality, in contrast of the RMSE, where the absence of lower bounds do not ensure the accomplishment of global optimality condition.

\section{Conclusion}
In this work, we have presented a formal analysis and empirical validation of the Information Teacher, a new data-driven decision rule designed to assess the optimality of regression models based on mutual information estimation. Theoretically, we have demonstrated that the Information Teacher exhibits strong convergence guarantees under the additive noise model assumption. % building upon the consistency framework proposed by González et al.~\cite{independence_detection} and extended by Ramírez et al.~\cite{ramirez}.

From an experimental perspective, we validate the effectiveness of the proposed metric in a controlled and a real-world regression setup. The synthetic scenario allowed for direct comparison against both the ground truth (oracle) error and the empirical MSE baseline, a widely used yet often misleading metric in model selection~\cite{bishop2006pattern}. Through a favorable and an unfavorable learning scenario, we showed that the Information Teacher is capable of discerning whether a regression model has attained the global minimum of the mean squared error (MSE), even in the absence of knowledge of the true model. In the favorable case, the Information Teacher converges to zero in alignment with the oracle error, effectively detecting optimality. In contrast, in the unfavorable case, the metric stabilizes above zero, reflecting suboptimal convergence due to model under-parameterization. The real-world scenario allowed us to confirm the capabilities of the Information Teacher to ensure the achievement of local optima in a low error regime, unlike the RMSE, where the absence of a lower bound does not give additional information about the performance.

These results confirm that the Information Teacher is not only theoretically grounded, but also practically informative in realistic conditions where the true model is unknown and the training data is limited or costly \cite{astro_data, ecommerce, healthcare}. In such contexts, the proposed approach offers a principled and data-driven criterion to assess algorithm performance without relying on inaccessible quantities.

Finally, by identifying the presence or absence of statistical dependence between inputs and residuals, the Information Teacher provides a solid basis for evaluating model adequacy, which is key for reliable deployment—especially in domains where interpretability matters~\cite{molnar2020interpretable}. As future work, this approach could be explored as a support tool for broader learning tasks such as meta-learning, early stopping, or model selection. Its ability to assess model behavior without requiring large validation sets makes it particularly appealing for scenarios with limited data or computational resources.

\section{Acknowledgements}
B. C., C. R., S. E., J. F. S., H. R. and M. O. acknowledge support from the Advanced Center for Electrical and Electronic Engineering (AC3E), funded by the Chilean ANID Basal Project (AFB240002). C. R.  was supported by ANID Subdirección de Capital Humano/Magíster-Nacional/2023-22230232 master's scholarship during this work. Additionally, J. F. S. and M. O. acknowledge support by grants of CONICYT-Chile Fondecyt Project 1250098. 

\bibliographystyle{IEEEtran}
\bibliography{ref}

\newpage
\onecolumn
\section{Supplementary Material}

\section{Proof of Theorem 2}

In this section, we prove the theoretical guarantees of the Information Teacher~---~Theorem 2. All results are framed under the additive noise model. 

Let $(X, Y) \sim \textup{add}(\eta, h; P_X)$ be a system governed by an additive noise model, where $\eta : \mathbb{R}^p \to \mathbb{R}^q$ is the true regression function and $h$ is the noise model. Consider now a measurable candidate regressor $\hat{\eta} : \mathbb{R}^p \to \mathbb{R}^q$, selected from a hypothesis space $\mathcal{F}$. As shown in the Main Manuscript, the central link between this setup and information-theoretic evaluation arises from the \emph{Model Equivalence Principle} (Theorem 1), which states that verifying whether $\hat{\eta}(X) \overset{\text{a.s.}}{=} \eta(X)$ is equivalent to testing the independence --via mutual information-- between the input $X$ and the residual $R = Y - \hat{\eta}(X)$.

The connection established by Theorem~1 allows us to understand the regression optimality assessment problem as a model drift detection problem; hence, some model drift properties presented in \cite{ramirez} can be adapted to our framework. In particular, Theorems 2, 3, and 4 from \cite{ramirez} address strong consistency, exponentially-fast detection of decision convergence on non-drifted systems and error vanishing to zero. The proof of our method properties in regard to the mentioned theorems is as follow:
\begin{enumerate}
    \item 
    Strong consistency --- We aim to establish the following convergence property of the data-driven Information Teacher: \begin{align}
        \eta(X) \overset{\textup{a.s.}}{=} \hat{\eta}(X) &\Rightarrow \hat{\mathbf{T}}(\hat{\eta};\mathcal{D}_m)\xrightarrow[m\rightarrow\infty]{\text{a.s.}}1,\\
        \eta(X) \overset{\textup{a.s.}}{\neq} \hat{\eta}(X) &\Rightarrow \hat{\mathbf{T}}(\hat{\eta};\mathcal{D}_m)\xrightarrow[m\rightarrow\infty]{\text{a.s.}}0,
\end{align}
The Information Teacher $\hat{\mathbf{T}}_m(\hat{\eta}, \mathcal{D}_m)$ acts as a test for independence between the input $X$ and the residual $R = Y - \hat{\eta}(X)$, based on $m$ i.i.d.\! samples $\mathcal{D}_m = \{(x_i, y_i)\}_{i=1}^m$. The core test statistic is the mutual information estimator $\hat{\mathcal{I}}_m(X; R)$, computed from the sample $Z_m = \{(x_i, r_i)\}_{i=1}^m$, where $r_i = y_i - \hat{\eta}(x_i)$.

The estimation process relies on a complexity-regularized binary tree partition of the space $\mathbb{R}^{p+q}$, constructed using axis-aligned splits at empirical medians \cite{tsp}. Each cell must contain at least $m \cdot b_m$ samples, and the regularization parameter $\lambda$ controls the structural complexity of the tree. Given a partition $\{A_l\}_{l \in S}$, the mutual information is estimated via
\begin{equation}
    \hat{\mathcal{I}}_m(X; R) = \sum_{l \in S} P_m(A_l) \cdot \log \left( \frac{P_m(A_l)}{P_m(A_l^{(1)} \times \mathbb{R}^q) \cdot P_m(\mathbb{R}^p \times A_l^{(2)})} \right),
\end{equation}
where $A_l = A_l^{(1)} \times A_l^{(2)}$ and $P_m$ denotes the empirical measures.
A binary rule $\psi^{\lambda,m}_{b_m, d_m, a_m} : \mathbb{R} \to \{0,1\}$, parameterized by $(\lambda, b_m, d_m, a_m)$, compares $\hat{\mathcal{I}}_m$ (parametrized implicitly by $\lambda$, $b_m$, and $d_m$) with the threshold $a_m$, deciding whether to reject that $X\perp R$ (i.e., \emph{almost-surely} equivalence between the student regressor $\hat{\eta}$ and the underlying encoder $\eta$). Accordingly, the data-driven information teacher can be written in terms of decision rule as:
\begin{equation}
\hat{\mathbf{T}}_m(\hat{\eta}; \mathcal{D}_m)\overset{\text{a.s.}}{=}  1 - \mathds{1}_{[a_m,\infty)}\left(\hat{\mathcal{I}}_m(X; R)\right).
\label{ITEq}
\end{equation}
Adopting the strong consistency definition (Def.~1, \cite{independence_detection}), the convergence behavior of $\hat{\mathbf{T}}_n$ is as follows:
\begin{align}
    ( X \perp Y - \hat{\eta}(X)) \Leftrightarrow (\mathbb{P} \left(\hat{\eta}(X) = \eta(X) \right) = 1 ) &\Rightarrow \left(\mathbb{P} \left( \lim_{n \to \infty} \hat{\mathbf{T}}_n(Z_n) = 1 \right) = 1\right), \\
    ( X \not\perp Y - \hat{\eta}(X)) \Leftrightarrow (\mathbb{P} \left(\hat{\eta}(X) = \eta(X) \right) < 1 ) &\Rightarrow \left(\mathbb{P} \left( \lim_{n \to \infty} \hat{\mathbf{T}}_n(Z_n) = 0 \right) = 1\right).
\end{align}
This behavior is equivalent to the existence of a binary sequence of decision rules $\psi^{\lambda,m}_{b_m, d_m, a_m}$ satisfying:
\begin{align}
    \left( X \perp Y - \hat{\eta}(X) \Rightarrow \mathbb{P} \left( \lim_{m \to \infty}\psi^{\lambda,m}_{b_m, d_m, a_m}(Z_m) = 0 \right) = 1 \right), \\
    \left( X \not\perp Y - \hat{\eta}(X) \Rightarrow \mathbb{P} \left( \lim_{m \to \infty}\psi^{\lambda,m}_{b_m, d_m, a_m}(Z_m) = 1 \right) = 1 \right).
\end{align}
Strong consistency for this family of decision schemes was formally established by (Theorem 3, \cite{independence_detection}), and further extended by (Theorem 2, \cite{ramirez}), under the following parameter conditions:
$$
b_m \approx m^{-\ell}, \ \ell \in (0,1/3), \ d_m \in \ell_1(\mathbb{N}), \quad 1/d_m = \mathcal{O}(\exp(m^{1/3})), \quad a_m \in o(1), \  \lambda\in \mathbb{R}^+.
$$
Given the equivalence $\hat{\eta}(X) \overset{\text{a.s.}}{=} \eta(X) \Leftrightarrow X \perp R$, this result confirms that the Information Teacher consistently identifies whether the student regressor $\hat{\eta}$ coincides with the true encoder $\eta$. This concludes the proof of the first part of Theorem 2. \hfill $\square$
 
    \item Exponentially-Fast Decision on the Global Optimum --- We now establish that, under the global optimum regime --i.e., $\eta(X)\overset{\text{a.s.}}{=}\hat{\eta}(X)$~-- the probability that the Information Teacher incorrectly rejects independence decays exponentially with the number of samples. Specifically, there exists a distribution-free constant $K > 0$ such that
\begin{equation}
    \mathbb{P}\left( \sup \left\{ k \in \mathbb{N} : \hat{\mathbf{T}}_k(\hat{\eta}, \mathcal{D}_k) = 0 \right\} \geq m \,\middle|\, \eta(X)\overset{\text{a.s.}}{=}\hat{\eta}(X) \right) \leq K \exp(-m^{1/3}).
    \label{eq:teacher_decay}
\end{equation}
To justify this result, we recall that under the parameter constraints
$$
b_m \approx m^{-\ell} \text{ with } \ell \in (0,1/3), \quad d_m\approx\exp(n^{-1/3}), \quad a_m \in o(1), \quad \lambda \in (0, \infty),
$$
Theorem 4 of \cite{independence_detection} and Theorem 3 of \cite{ramirez} guarantee that for any decision scheme $\Psi=(\psi^{\lambda,m}_{b_m, d_m, a_m}
)_{m \in \mathbb{N}}$ based on the mutual information estimation introduced by \cite{tsp}, under independence, the probability that the induced tree-structured partition remains non-trivial (which implies some statistical dependency is detected) beyond $m$ samples decays exponentially. This is,
\begin{equation}
    \mathbb{P}\left( \mathcal{T}_0^{\Psi}\left( (Z_n)_{n \geq 1} \right) \geq m \right) \leq K \exp(-m^{1/3}),
    \label{eq:exp_decay}
\end{equation}
where $\mathcal{T}_0^{\Psi}$ denotes the latest time the decision rule deviates from outputting $0$, which corresponds to the moment independence is always further detected.

Since the Information Teacher $\hat{\mathbf{T}}_m$ is defined via a threshold on the same mutual information estimator used in the decision scheme $\Psi$, the equivalence
$$
\hat{\mathbf{T}}_k(\hat{\eta}, \mathcal{D}_k) = 0 \quad \Leftrightarrow \quad \text{independence not detected with $k$ samples}
$$
holds for all $k \in \mathbb{N}$. Therefore, Eq. \eqref{eq:teacher_decay} follows directly from \eqref{eq:exp_decay}, and completes the second part of the proof of Theorem 2. \hfill $\square$

     \item Error Convergence to Zero --- We now address the convergence of the type I and type II error probabilities associated with the Information Teacher. As shown in (Theorem 4, \cite{ramirez}), the mutual information-based decision rule exhibits strong non-asymptotic guarantees: the significance level decays exponentially in the number of samples, while the statistical power converges to one. Specifically, for the decision rule $\psi^{\lambda, m}_{b_m, d_m, a_m}$, we have the following bounds for all $m \in \mathbb{N}$ and some universal constant $K > 0$:
\begin{align}
    \alpha_{\psi_m} &= \mathbb{P} \left( \psi^{\lambda,m}_{b_m, d_m, a_m}(Z_m) = 1 \mid X\perp X-\hat{\eta}(X) \right) \leq K \cdot \exp(-m^{1/3}), \label{eq:alpha_m} \\
    \lim_{m \to \infty} (1 - \beta_{\psi_m}) &= 1, \quad \text{where } \beta_{\psi_m} = \mathbb{P} \left( \psi^{\lambda,m}_{b_m, d_m, a_m}(Z_m) = 0 \mid X\not\perp X-\hat{\eta}(X) \right). \label{eq:beta_m}
\end{align}
Given that the data-driven Information Teacher $\hat{\mathbf{T}}_m(\hat{\eta}, \mathcal{D}_m)$ is defined as a thresholded version of the same mutual information estimator used in $\psi^{\lambda, m}_{b_m, d_m, a_m}$, these guarantees extend to the Information Teacher. Therefore, for some constant $\Tilde{K} > 0$, we conclude:
\begin{align}
    \widetilde{\alpha}_m &= \mathbb{P} \left( \hat{\mathbf{T}}_m(\hat{\eta}; \mathcal{D}_m) = 0 \mid \eta(X)\overset{\text{a.s.}}{=}\hat{\eta}(X) \right) \leq \Tilde{K} \cdot \exp(-m^{1/3}), \label{eq:alpha_tilde_m} \\
    \lim_{m \to \infty} (1 - \widetilde{\beta}_m) &= 1, \quad \text{where } \widetilde{\beta}_m = \mathbb{P} \left( \hat{\mathbf{T}}_m(\hat{\eta}; \mathcal{D}_m) = 1 \mid \eta(X)\overset{\text{a.s.}}{\neq}\hat{\eta}(X) \right). \label{eq:beta_tilde_m}
\end{align}
This confirms that the Information Teacher retains strong finite-sample performance, with exponentially vanishing significance and asymptotically perfect detection power. \hfill $\square$

\end{enumerate}
\end{document}